\documentclass[journal,twoside]{IEEEtran}

\usepackage{amsmath}
\usepackage{hyperref}
\usepackage{graphicx}
\usepackage[export]{adjustbox}
\usepackage{siunitx}

\makeatletter
\let\NAT@parse\undefined
\makeatother

\usepackage{multirow}
\usepackage{bbm}
\usepackage{balance}
\usepackage{bm}
\usepackage[numbers,sort&compress]{natbib}
\usepackage{booktabs}
\usepackage{soul}

\ifCLASSOPTIONcompsoc
  \usepackage[caption=false,font=normalsize,labelfont=sf,textfont=sf]{subfig}
\else
  \usepackage[caption=false,font=footnotesize]{subfig}
\fi

\begin{document}

\title{Graph Embedding Augmented\\Skill Rating System}

\author{Jiasheng Wang
\thanks{
Manuscript received November 17, 2021; revised May 30, 2022 and September 26, 2022; accepted October 18, 2022; date of the current version November 9, 2022. 
}
\thanks{
Jiasheng Wang is with the Department of Statistics, Shanghai University, Shanghai 200444, China (e-mail: wangjiasheng@shu.edu.cn).
}
}
\markboth{IEEE TRANSACTIONS ON GAMES,~Vol.~XX,~No.~XX,~MONTH~20XX}%
{Wang: Graph Embedding Augmented Skill Rating System}

\maketitle

\begin{abstract}

This paper presents a framework for learning player embeddings in competitive games and events. 
Players and their win-loss relationships are modeled as a skill gap graph, which is an undirected weighted graph. 
The player embeddings are learned from the graph using a random walk-based graph embedding method and can reflect the relative skill levels among players. 
Embeddings are low-dimensional vector representations that can be conveniently applied to subsequent tasks while still preserving the topological relationships in a graph. 
In the latter part of this paper, Graphical Elo (GElo) is introduced as an application of player embeddings when rating player skills. 
GElo is an extension of the classic Elo rating system. 
It constructs a skill gap graph based on player match histories and learns player embeddings from it. 
Afterward, the rating scores that were calculated by Elo are adjusted according to player activeness and cosine similarities among player embeddings. 
GElo can be executed offline and in parallel, and it is non-intrusive to existing rating systems. 
Experiments on public datasets show that GElo makes a more reliable evaluation of player skill levels than vanilla Elo. 
The experimental results suggest potential applications of player embeddings in competitive games and events. 

\end{abstract}


\IEEEpeerreviewmaketitle


\section{Introduction}

\IEEEPARstart{S}{kill} rating systems are widely applied in many competitive games and events. 
Famous skill rating systems such as Elo, Glicko, TrueSkill~\cite{elo1978rating, glickman1995comprehensive,glickman2012example, herbrich2006trueskill} usually represent a player with a single number (i.e., rating score) to describe skill level. 
The rating score of a player is continually updated depending on the match results. 
After each match, the winning player subtracts rating scores from the losing player. 
The subtracted value is decided generally as follows: 
\begin{enumerate}
    \item if the higher-rated player wins, a few points are taken from the lower-rated player;
    \item if the lower-rated player wins, a lot of points are taken from the higher-rated player.
\end{enumerate}

Initially designed for chess games, skill rating systems nowadays have a much wider range of application domains, including various card \& board games, sports leagues, and online competitive video games. 
One common application is matchmaking in online competitive video games such as Counter-Strike. 
With a properly designed skill rating system, we can efficiently identify players with similar skill levels and match them together, creating evenly contested matches. 
When a novice is pitted against an experienced opponent, neither is likely to enjoy the match. 
The rating scores calculated by rating systems can also be used for a variety of purposes. 
For example, the rating scores can act as a qualification for high-level tournaments or be displayed to players and audiences to stimulate their interests and competitions~\cite{herbrich2006trueskill}. 

Due to the importance of the skill rating systems, many efforts have been made to improve their evaluation performance. 
Some works modify rating rules or design new ones. 
Some works incorporate feature engineering techniques to exploit in-game performance of players. 
Some works consider historical information. 
These efforts provide meaningful insights into the domain. 

Investigations into the relations and interactions among players may also be beneficial for the skill rating process. 
Player skills are inferred from their winning and losing records against others, so the primary relationships in competitive events are the win-loss relationships.
However, how to properly describe and handle the win-loss relationships is a challenge we face. 
Therefore, in this paper, we propose a framework that models the win-loss relationships among players as a \textit{skill gap graph} $G \left(V, E \right)$ and learns player embeddings from this graph. 
Embeddings are low-dimensional vector representations that can be conveniently applied to downstream tasks while still keeping the relational information. 
We present Graphical Elo (GElo) as an application of player embeddings. 
GElo is an extension of the classic Elo rating system that incorporates player embeddings to help make better evaluations of player skill levels. 

We conduct comprehensive experiments to test GElo using public datasets from online competitive video games and sports leagues.
Experiments show that the skill evaluation performance of GElo exceeds vanilla Elo in four of five datasets. 
The experimental results also suggest potential applications of player embeddings in competitive games and events, as they preserve the information about the win-loss relationships among players.

The main contributions of this paper include the following:
\begin{itemize}
  \item a framework for learning player embeddings;  
  \item definitions of \textit{skill gap graph} and its interaction patterns; 
  \item introducing Graphical Elo, an extension of Elo rating system that utilizes player embedding. 
\end{itemize}

The remaining part of this paper proceeds as follows: 
Section~II reviews previous research on graph embedding and skill rating systems. 
Section~III delves into the definition of the skill gap graph and its interaction patterns, followed by an overview of GElo. 
In Section~IV, we discuss the method of GElo in detail. 
Section~V conducts comprehensive experiments to examine the overall performance of GElo. 
We conclude this paper in Section~VI by reviewing the contributions and limitations of the work, as well as outlining future works.


\section{Backgrounds}

\subsection{Graph Embedding}

Graph is a mathematical structure that can model pairwise relations between objects in many scenes, including social sciences, biology, linguistics~\cite{goyal2018graph}, etc.

We want to learn graph embedding to utilize the abundant while complex information of the graph. 
Our goal is to find the mapping function $\phi: V \mapsto \mathbbm{R}^{|V| \times d}$ that converts the graph into vector representations in a low-dimensional Euclidean space, where $|V|$ is object size, and $d$ is the dimension of embedding vectors. 
The vectorized graph embedding enables us to perform extensive tasks with ease while still preserving the topological relationships in the graph. 
These advantages make graph embedding widely adopted in recommendation systems and knowledge graph domains. 

With the development of deep learning, there are emerging deep learning-based graph embedding methods. 
One type of which employs a random walk approach. 
DeepWalk~\cite{perozzi2014deepwalk}, inspired by the word embedding method word2vec~\cite{mikolov2013efficient}, learns node embeddings from a collection of random walk sequences using neural networks. 
Each random walk sequence sampled from the graph corresponds to a sentence from the text corpus, where a node corresponds to a word. 
Then, Skip-gram model~\cite{mikolov2013distributed} is applied to the random walk sequences to maximize the likelihood of observing the neighborhood of a node, given the current node embedding~\cite{cai2018comprehensive}. 
node2vec~\cite{grover2016node2vec} improves DeepWalk by featuring a biased second-order random walk on weighted graph. 
Other widely applied deep learning-based graph embedding methods include LINE~\cite{tang2015line}, SDNE~\cite{wang2016structural}, and the family of Graph Neural Networks (GNNs) models. 

In this paper, we use the random walk-based graph embedding method inspired by DeepWalk to learn node representations.
We assume that nodes being similar in the graph tend to have similar representations, which is an assumption also adopted by many works in this domain~\cite{perozzi2014deepwalk, grover2016node2vec, wang2016structural}.

\subsection{Skill Rating System}

Skill rating systems usually use a single scalar to denote player skill levels, such as Elo, Glicko, TrueSkill~\cite{elo1978rating, glickman1995comprehensive, glickman2012example, herbrich2006trueskill}, and their extensions~\cite{kovalchik2020extension, huang2008ranking, zhang2010factor, weng2011bayesian, delong2011teamskill, makhijani2019parametric, ebtekar2021elo}. 
They generally derive rating scores from the win-loss records of players. 

Many efforts have been made on skill rating systems to improve their skill evaluation performance. 
Some works develop these rating systems by emphasizing individual skills. 
They incorporate additional player information by feature engineering on the in-game performance of players with expert knowledge~\cite{delalleau2012beyond, chen2016predicting, minka2018trueskill}. 
Some works consider the historical information of players. 
Works that analyze the historical information to get more reliable predictions include Edo~\cite{hunter2004mm} and WHR~\cite{coulom2008whole}. 
Four kinds of rating systems have been summarized regarding the handling method of historical player data: Static Rating Systems, Incremental Rating Systems, Decayed-history Rating Systems, and Accurate Bayesian Inference~\cite{coulom2008whole}. 
Competitive games usually include team vs. team scenarios, so some works~\cite{semenov2016performance, li2018learning, sapienza2019deep} look into the cooperative effect of team composition. 
These works provide meaningful insights into the domain. 

Competitive games and events are full of engagements and interactions among players. 
Since graph is a powerful tool to measure such inter-relationships, some works investigate the skill rating process from a graphical perspective by graph spectral theory method~\cite{katz1953new, park2005network, lasek2013predictive, Ricatte2020SkillRF}. 
Works interpreting the skill rating process by the deep learning-based graph embedding method include OptMatch~\cite{gong2020optmatch}.
Based on player match records, two types of relations are concluded to represent the interplays among players, i.e., 1) synergy relation built from the win-lose relationships, and 2) suppression relation that indicates the advantage of one over another. 
However, these works mainly focus on the matchmaking process or the cooperation effects for players. 
We mainly focus on calculating the rating scores of players like the classic rating systems (e.g., Elo, Glicko, Trueskill, etc.). 
Our work also features a detailed discussion of the skill gap graph.


\section{Skill Gap Graph}

\subsection{Graphical Model of Skill Rating Process}

In competitive events, we acknowledge player skills mainly by comparing the winning and losing records among players. 
When two players compete, the player who wins more games than the other is considered stronger. 
These win-loss relations among players constitute a network, from which we can infer player skill levels by comparing any of the players. 
Even for players who have not directly engaged, we can still infer their skills indirectly according to their win-loss relationships with other players. 
The win-loss records are typical relationships in competitive events, which contain information that can help us acknowledge the relative skill levels among players. 
Hence, the skill rating process is an investigation of the win-loss relations among players by its very nature. 
If we can properly model and exploit such inter-relationships, it shall benefit our evaluation of the player skills. 

We use a graph to describe and model the win-loss relationships among players. 
Based on player match histories, we construct a $\textrm{skill gap graph}$, which is an undirected weighted graph $G \left(V, E \right)$. 
Each node $v_i \in V$ represents a player $p_i$. 
An edge $e_{ij}$ connecting $v_i$ and $v_j$ indicates the engagements between $p_i$ and $p_j$ (i.e., matches have taken place between them). 
Then, we define the edge weight $w_{ij}$ as the \textit{skill gap} between $p_i$ and $p_j$. 
The term ``skill gap'' refers to the difference in skill levels, which is measured by the win-loss records between a pair of players. 
We emphasize ``skill gap'' instead of ``skill level similarity'' because the primary goal when modeling the relationships is maximizing the difference in player skill levels. 
Only through sharp contrasts among player skills can we quickly and clearly identify the relative skill levels among players. 
Skill gaps among players are the basic and crucial relationships that form a skill gap graph. 

Skill gap graph is the core concept of our graphical skill rating model. 
It captures the win-loss relationships among players in a topological form, enabling us to interpret the skill rating process from a graphical perspective.

\subsection{Convert Graph into Vector Spaces}

\begin{figure}[!t]
    \centering 
    \includegraphics[height=3.43cm]{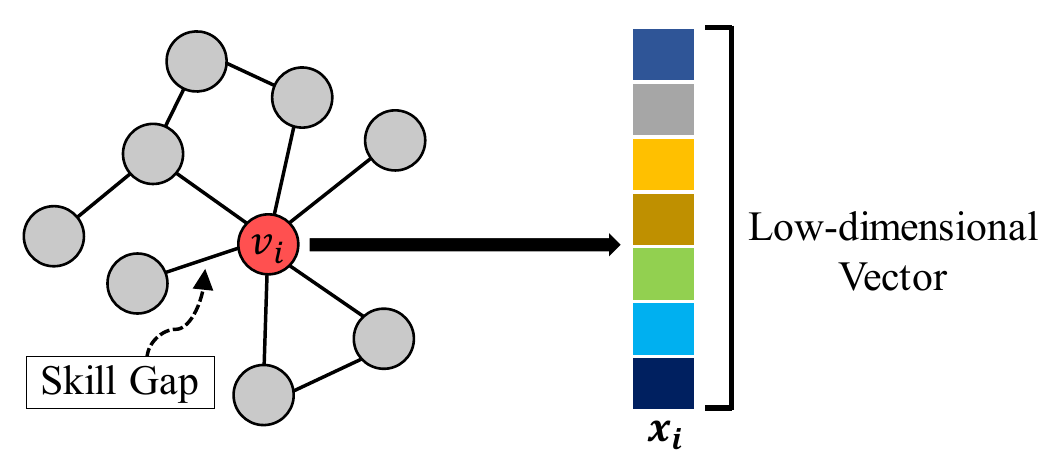} 
    \caption{
    Schematic of learning node (player) embeddings. 
    Skill gap graph is an undirected weighted graph. 
    The edge weight is the skill gap between a pair of players. 
    Nodes from the original graph are in a high-dimensional irregular form. 
    Graph embedding method can map the nodes into a low-dimensional vector space while still keeping the topological relationships. 
    The obtained vector representations can be easily used for subsequent tasks. 
    } 
    \label{fig:graph2vec} 
\end{figure}

Although graph is a powerful tool for modeling relationships, it is difficult to deal with the graph data directly. 
To handle the abundant while complex information of the skill gap graph, a graph embedding method is necessary. 
The graph embedding method aims to find the mapping $\phi: V \mapsto \mathbbm{R}^{|V| \times d}$ to get embeddings from a graph $G \left(V, E \right)$. 
As illustrated in \autoref{fig:graph2vec}, the graph embedding method maps nodes into vectors of real numbers, allowing applications to various tasks~\cite{xu2021understanding}. 

With the assumption that nodes being similar in a graph (i.e., players who have small skill gaps with each other) tend to have similar embeddings in vector space, the skill rating process can then be viewed as a process of identifying nodes (players) with similar skill levels and grouping them together. 
That is, we are inferring and ranking the relative skill levels of players regarding their representational similarities. 

We employ a random walk-based graph embedding method inspired by DeepWalk~\cite{perozzi2014deepwalk}, which learns node embeddings by modeling a stream of random walk sequences using neural networks. 
It can be parallelized with ease and is more computationally efficient than advanced models such as GNNs. 

\subsection{Interaction Patterns}

Interaction patterns explain how the skill rating process works on a skill gap graph. 
We summarize three categories of the interaction patterns: 1)~\textit{homophily}, 2)~\textit{structural equivalence}, and 3)~\textit{deduction}. 
Here we use a toy example in \autoref{fig:toy_example} to illustrate these interaction patterns. 

In \autoref{fig:toy_example}, there are six nodes $\{a, b, c, d, e, f\}$
that represent six players. 
Each edge can be observed in two forms: a solid line representing a small skill gap, or a dotted line representing a large skill gap. 
Different combinations of nodes exhibit various kinds of interaction patterns: 

\subsubsection{Homophily}

Homophily looks at local proximity. 
If two nodes have a small skill gap between them, we can infer that they have a similar skill level. 
Nodes $\{d, e\}$ in \autoref{fig:toy_example} is one example of the homophily.

\subsubsection{Structural Equivalence}

Structural equivalence captures a situation where some nodes share similar relationships with other nodes. 
Instead of concentrating on the direct connectivity between nodes as homophily does, structural equivalence emphasizes the neighboring structures of nodes. 
Nodes in a network can be far apart while still playing the same structural role. 
In \autoref{fig:toy_example}, nodes \{$b, f$\} have similar neighboring structures because both of them have small skill gaps to $a$ and large skill gaps to $c$, so they probably have similar skill levels.

\subsubsection{Deduction}

\begin{figure}[!t]
    \centering
    \includegraphics[height=3.1cm]{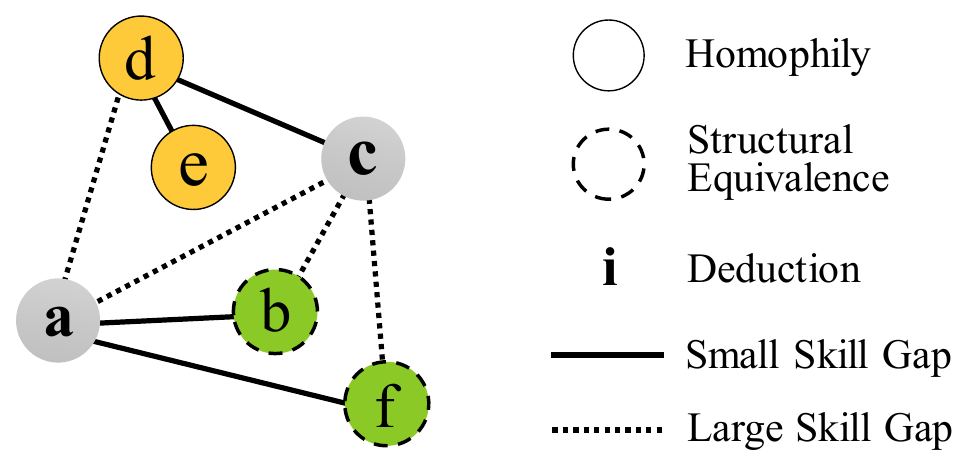}
    \caption{
    Metaphorical illustration of skill gap graph. 
    We classify the interaction patterns into three categories: homophily, structural equivalence, and deduction. 
    Nodes $\{d, e\}$ reflect a homophily. 
    Nodes $\{b, f\}$ reflect a structural equivalence. 
    Nodes $\{a, c\}$ in triangle groups $\{a, b, c\}$ and $\{a, d, c\}$ play a role of deduction. 
    }
    \label{fig:toy_example}
\end{figure}

Deduction, or deductive reasoning, compares skill levels regarding reciprocal relationships in a large group. 
In \autoref{fig:toy_example}, nodes $\{a, c\}$ are members of triangle node groups $\{a, b, c\}$ and $\{a, d, c\}$.
From these two triangle groups, we can acknowledge that $a \approx b \not\approx c$ and $a \not\approx d \approx c$ in skill level. 
With the known skill level relation $a \not\approx c$, we can then infer that $P(b \not \approx d) > P(b \approx d)$ (e.g., if $a$ is \textit{slightly weaker} than $b$, and meanwhile $a$ is \textit{greatly weaker} than $d$, then $b$ has a possibility of being \textit{slightly weaker} than $d$. 
This is a situation where $b \approx d$ and its possibility is relatively lower because $a \not\approx c$). 
Though $b$ and $d$ are not directly connected, we can still deduce their relationship through $\{a, c\}$. 

\subsection{Overview of Graphical Elo}

Graphical Elo (GElo) is an extension of the classic Elo rating system~\cite{elo1978rating}. 
It constructs a skill gap graph based on player match histories and learns player embeddings from the graph. 
Afterward, the player embeddings can be used in a variety of ways. 
In our implementation of GElo, we will adjust the rating scores according to player activeness and cosine similarities among the player embeddings. 

We relieve the asynchronism of Elo.
For inactive players who have only played a few matches, their rating scores will fluctuate around the initial value. 
This sometimes leads to a weird situation: though an active player (a player who has played many matches) has a higher skill level than an inactive player, the rating score of this active player may still be lower than that of the inactive player, because the inactive player has not played enough matches to receive a reliable rating score. 
Actually, the lower-rated active players have potentially better skills because they have more game experience. 
The asynchronism of the rating systems impacts the reliability of the rating scores to some extent. 
So, a post-adjustment of the rating scores may relieve this issue. 
The adjustment values are scaled by the cosine similarities among the player embeddings. 

GElo is non-intrusive and rather acts as a supplement to base rating systems. 
GElo can be executed offline and in parallel. 
Same as vanilla Elo, in-game information such as player performances is not necessary. 
These advantages show the ease of deployment of GElo.


\section{Method of GElo}

\subsection{Elo Rating System}

Elo rating system~\cite{elo1978rating} is the base system of GElo. 
It uses a single number to describe the skill level of a player. 
Initially, a default rating score of $1500$ is assigned to every player. 
The player rating score will be continuously updated in terms of the match results. 

Let $p_i$ denote player $i$, and $r_i$ denote the rating score of $p_i$.
The updated value of the rating score after a certain match is determined by the difference between the actual match result $s_i$ and the expected win rate $\mathring{s}_i$. 
For a match between $p_a$ and $p_b$, if $p_a$ wins against $p_b$, then their actual match results are $s_a=1$, $s_b=0$, otherwise $s_a=0$, $s_b=1$. 
The expected win rates $\mathring{s}_a$ and $\mathring{s}_b$ are estimated according to the score difference in $r_a$ and $r_b$:
\begin{equation}
\label{eqn:exp_win_rate}
    \mathring{s}_a = 1 - \mathring{s}_b = \dfrac{1}{1+10^{-(r_a - r_b)/400}}.
\end{equation}
Equation \eqref{eqn:exp_win_rate} uses an S-shaped curve within the output range of $(0, 1)$. A difference of $200$ in $\lvert r_a - r_b \rvert$ will make the higher-rated player have an expected win rate of approximately $0.76$~\cite{elo1978rating}. 
Then, the new rating scores $\hat{r_{a}}$ and $\hat{r_{b}}$ after the match for both players are updated as follows:

\begin{alignat*}{2}
    \hat{r_{a}} &= r_a + k \times (s_a - \mathring{s}_a& );\\
    \hat{r_{b}} &= r_b + k \times (s_b - \mathring{s}_b& ). 
\end{alignat*}
K-factor $k$ is a user-defined positive constant that scales the updating values. 

To make Elo feasible for team vs. team competitions, in a match we simply view two teams as two ``players''. 
The rating score of a team is the average rating score of all team members. 
After a match, we update scores for every single player using Elo algorithm on each player separately. 
To be specific, one of the $r_a$ and $r_b$ in \eqref{eqn:exp_win_rate} is the rating score of a certain player, while the other is the average rating score of the opposite team.

\subsection{Graph Construction}

We build a skill gap graph $G = (V, E)$ based on player match histories, where $G$ is an undirected weighted graph.
Node $v_i \in V$ represents player $p_i$. 
An edge $e_{ab} \in E$ will connect $v_a$ and $v_b$ if match(es) happened between $p_a$ and $p_b$. 

$w_{ab}$ is the edge weight of $e_{ab}$, which describes the skill gap between $p_a$ and $p_b$. 
The skill gap is measured by the win-loss results and the number of matches between the pair of players. 
Let $o_{ab}^n$ denote the result of the $n$th match between $p_a$ and $p_b$, then two possible results of $o_{ab}^{n}$ include: 
\begin{equation*}
    o_{ab}^{n}=
    \begin{cases}
    +1,\qquad\text{if $a$ wins against $b$ in the $n$th match;} \\ 
    -1,\qquad\text{if $a$ loses to $b$ in the $n$th match.}
    \end{cases} 
\end{equation*}
The total number of matches that happened between $p_a$ and $p_b$ is denoted as $m_{ab}$. 

Afterward, we can estimate the skill gap from $o_{ab}^n$ and $m_{ab}$. 
The edge weight $w_{ab}$, which describes the skill gap, is then defined as follows: 
\begin{gather*}
    w_{ab}=
    \begin{cases}
    1-\tanh\left( \dfrac{\lvert \sum_{n=1}^{m_{a b}}{o_{a b}^{n} \rvert}}{m_{ab}} \right),\qquad &\text{if $m_{ab} \geq 2$;}\\[11pt]
    0.01,\qquad &\text{if $m_{ab} = 1$}. 
    \end{cases}
\end{gather*}

When $m_{ab} \geq 2$, as $\lvert \sum_{n=1}^{m_{a b}}{o_{a b}^{n}} \rvert$ increases (normalized by $m_{ab}$), the skill gap becomes larger and will output a small $w_{ab}$. 
In the case of $m_{ab} = 1$, however, it is not reliable to judge the skill gap. 
The win rate is pointless when there is only one match. 
For this reason, we set $w_{ab} = 0.01$, which is a constant far smaller than the minimum of $w_{ab}$ given $m_{ab} \geq 2$ (approximately $0.238$).
Therefore, during the following graph embedding procedure, random walkers will be less likely to travel through the edges that record only one match, thus alleviating the impact of such unreliable information. 

\subsection{Learn Player Representations}

Random walk-based graph embedding method is used to learn player embeddings~\cite{perozzi2014deepwalk}. 
It learns node embeddings using neural networks from a collection of random walk sequences based on Skip-gram model~\cite{mikolov2013distributed}. 

First, we will perform random walks on the skill gap graph to generate node sequences as the training data corpus.
Starting from each node in the graph, the random walker will continuously transit to neighboring nodes (i.e., nodes that are directly connected) and terminate until the sequence length reaches a given limit. 
Supposing a random walker arrives at $v_i$, then the next node $v_t$ it will transit to will be a neighbor of $v_i$, $e_{ti} \in E $. 
Usually, a node has many neighboring nodes, so to which neighbor $v_t$ the random walker will transit is selected proportional to the edge weights: 
\begin{equation*}
    P(v_{t} \mid v_{i})=\dfrac{w_{ti}}{\sum w_{*i}},
\end{equation*}
where $\sum w_{*i}$ is the sum weight of all edges that connect $v_i$. 

After generating enough random walk sequences, our next goal is to transform the corpus of node sequences into vector representations. 
Let $\phi: V \mapsto \mathbbm{R}^{|V| \times d}$ be the mapping function from a graph to vectorized representations, where $|V|$ is the player amounts and $d$ is the dimension of the vector embeddings. 
Then, the following function is the optimizing objective of neural networks, which maximizes the log probability of observing a neighborhood for a node $v_i$ conditioned on its vectorized representation given by $\phi$:
\begin{equation}
\label{eqn:obj}
    \max_{\phi}{\sum_{v_i \in V}{\log P(N(v_i) \mid \phi(v_i))}},
\end{equation}
where $N(v_i)$ is the neighborhood of $v_i$ with a context size $u$. 
The neighborhood is that, for example, suppose there is a node sequence $[v_1, v_2, v_3, \dots, v_6]$, then $N(v_2)$ with $u = 1$ includes $\{v_1, v_3\}$; $N(v_4)$ with $u = 2$ includes $\{v_2, v_3, v_5, v_6\}$. 

To make the optimization objective in \eqref{eqn:obj} tractable, we give two assumptions~\cite{grover2016node2vec}: 1)~\textit{Conditional Independence} assumption and 2)~\textit{Symmetric Feature Space} assumption: 

\subsubsection{Conditional Independence}
We assume neighbors are independent of each other inside a neighborhood. 
Let $n_i \in N(v_i)$ be one of the neighboring nodes of $v_i$. 
Then, given the representation of the source node $\phi(v_i)$, the likelihood of observing a neighbor $n_i$ is independent of observing any other neighboring node: 
\begin{equation}
\label{eqn:ind}
    P(N(v_i) \mid \phi (v_i)) = \prod_{n_i \in N(v_i)}{P(n_i \mid \phi(v_i))};
\end{equation}

\subsubsection{Symmetric Feature Space}
We assume a source node and a neighbor node have a symmetric effect on each other in vector space. The conditional likelihood of every source-neighbor node pair $\phi(v_i)$ and $n_i$ is modeled as a softmax unit parameterized by a dot product of their representations: 
\begin{equation}
\label{eqn:softmax}
    P(n_i \mid \phi(v_i)) = \dfrac{\exp(\phi(n_i) \cdot \phi(v_i))}{\sum_{v \in V}{\exp(\phi (v) \cdot \phi(v_i))}}. 
\end{equation}

With the above two assumptions \eqref{eqn:ind} and \eqref{eqn:softmax}, we can now simplify the objective in \eqref{eqn:obj} to:
\begin{align}
    &\phantom{{}={}}\sum_{v_i \in V}{\log P(N(v_i) \mid \phi(v_i))} \nonumber \\
    &=\sum_{v_i \in V}{\log \prod_{n_i \in N(v_i)}{P(n_i \mid \phi(v_i))}} \nonumber \\
    &=\sum_{v_i \in V}{\sum_{n_i \in N(v_i)}{\log \left( \dfrac{\exp(\phi(n_i) \cdot \phi(v_i))}{\sum_{v \in V}{\exp(\phi(v) \cdot \phi(v_i))}} \right) }}. \nonumber
\end{align}
Let $Z_{v_i}=\sum_{v\in V} \exp(\phi(v) \cdot \phi(v_i))$, then our optimization objective is as follows: 
\begin{equation}
\label{eqn:obj_simple}
    \max_{\phi}{\sum_{v_i \in V}{\left[ \sum_{n_i \in N(v_i)}{\phi(n_i) \cdot \phi(v_i)-\lvert N(v_i)\rvert \log(Z_{v_i})}\right]}}. 
\end{equation}
$Z_{v_i}$ can be approximated using negative sampling~\cite{mikolov2013distributed}. 

Equation~\eqref{eqn:obj_simple} can be optimized using stochastic gradient ascent. 
The player embeddings $\bm{x}_i \in \bm{X}$ are extracted from the hidden layer of the neural network.

\subsection{Post-adjustment of Player Rating Scores}

The player embeddings preserve the relationships among players involving their win-loss records, and their vectorized representations enable them to be easily applied to various tasks. 
In our implementation of GElo, we feature a post-adjustment process of the rating scores to alleviate the asynchronism of vanilla Elo. 
Low-rated active players may have better skills than high-rated inactive players, so these low-rated active players will be granted bonus points upon their rating scores to set their positions right. 
The bonus points are scaled by the cosine similarities calculated using $\bm{x}_i \in \bm{X}$. 

\begin{figure}[!t]
    \centering 
    \includegraphics[height=4.11cm]{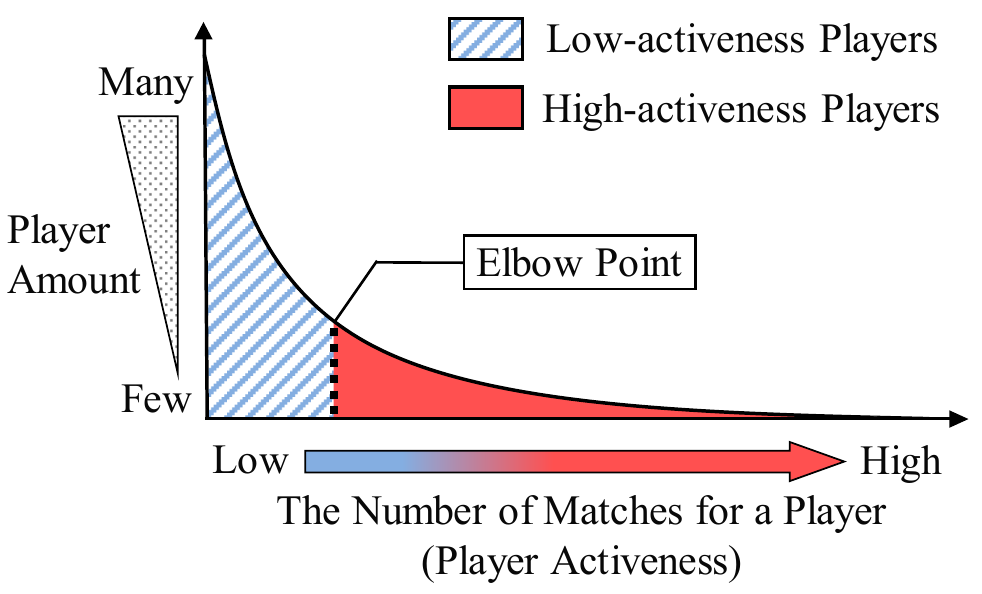}
    \caption{
    In terms of player activeness, the player amount follows a long-tailed distribution. 
    Player activeness is judged by the number of matches a player has played in a period. 
    As the activeness increases, the number of corresponding players will decrease drastically. 
    We would like to find the number of matches that can properly classify players as active or inactive.  
    The optimal threshold is the elbow point, which usually has the maximum second derivative. 
    }
    \label{fig:long_tailed}
\end{figure}

We provide a heuristic approach to quickly identify the active players. 
As a common situation depicted in \autoref{fig:long_tailed}, the player amount follows a long-tailed distribution in terms of their playing activeness. 
Active players consist of the tail part on the right, while inactive players congregate in the head part on the left. 
We would like to locate the elbow point of the distribution curve. 
Intuitively, the elbow point can properly divide the player group into active and inactive players. 

The elbow point is usually where the maximum second derivative is. 
Let $\#M(i)$ denote the number of players who play $i$ matches. 
According to Taylor's theorem, the second derivative $\#M^{\prime\prime}(i)$ can be approximated as:
\begin{equation*}
    \#M^{\prime\prime}(i) = \#M(i+1) + \#M(i-1)-2 \times \#M(i).
\end{equation*}
The $i$ that maximizes $\#M^{\prime\prime}(i)$ is the elbow point. 
Players who have played no less than $i$ matches are considered active. 

Among the active players, we first choose a top player (i.e., the player with the highest rating score) and a bottom player (i.e., the player with the lowest rating score). 
They are regarded as benchmarks when deciding bonus points, and their embeddings are denoted as $\bm{x}_\textrm{top}$ and $\bm{x}_\textrm{btm}$. 
Then, we will decide bonus points for all active players regarding their respective skill gaps to the top player. 
The skill gaps here, unlike the ones given by edge weights in the skill gap graph, are measured in vector space using player embeddings $\bm{x}_i \in \bm{X}$, which keep the topological information of the skill gap graph and enable direct comparison between any players, even if they have no direct engagement. 
A player with a small skill gap to the top player is considered strong, and thus will receive more bonus points. 

For two players $p_a$ and $p_b$, the skill gap between them in vector space can be derived from the cosine similarity between their embeddings $\bm{x}_a$ and $\bm{x}_b$, which is as follows: 
\begin{equation*}
\label{eqn:avg_cosm}
    \textrm{cosm}(a,b)=\left\lvert \dfrac{\bm{x}_a \cdot \bm{x}_b}{\lVert \bm{x}_a \rVert \lVert \bm{x}_b \rVert} \right\rvert.
\end{equation*}
$\textrm{cosm}(a,b)$ determines whether $\bm{x}_a$ and $\bm{x}_b$ are pointing in roughly the same direction by measuring the angle between them in vector space. 
A large $\textrm{cosm}(a,b)$ means $\bm{x}_a$ and $\bm{x}_b$ are close, and thus the skill gap between $p_a$ and $p_b$ is small.

We define a function that calculates the average cosine similarity between $\bm{x}_i$ and $\bm{x}_\textrm{top}$ as follows: 
\begin{equation*}
    \overline{\textrm{cosm}}(i,\textrm{top}) = \dfrac{\textrm{cosm}(i,\textrm{top})+(1-\textrm{cosm}(i,\textrm{btm}))}{2},
\end{equation*}
which is the average of
1)~direct cosine similarity to the top player $\textrm{cosm}(i,\textrm{top})$, and 2)~reverse cosine similarity to the bottom player $1-\textrm{cosm}(i,\textrm{btm})$. 
Doing so yields a more robust result than solely depending on one of 1) and 2). 

\begin{figure}
    \centering
    \includegraphics[height=3.15cm]{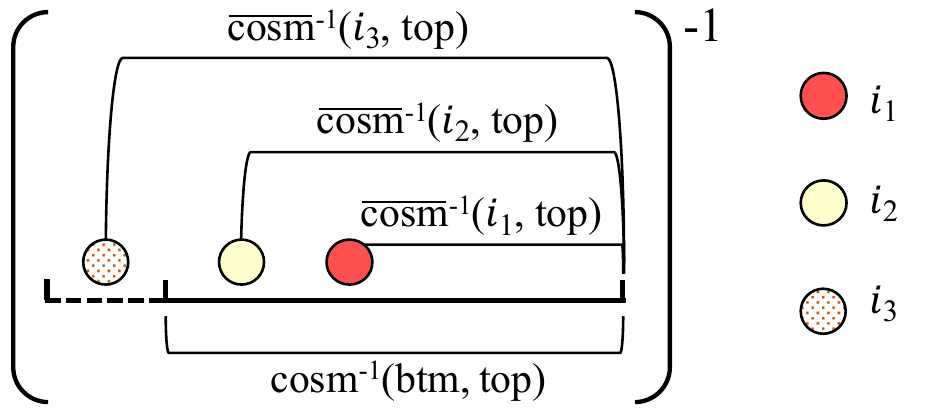}
    \caption{
    A skill gap perspective (in vector space) to explain $\textrm{Sim}_\textrm{top}(i)$. 
    Function $\overline{\textrm{cosm}}^{-1}(i,\textrm{top})$ represents the skill gap to the top player for $p_i$. 
    Since $\overline{\textrm{cosm}}^{-1}(i_1,\textrm{top}) < \overline{\textrm{cosm}}^{-1}(i_2,\textrm{top})$, we consider $p_{i_1}$ is stronger than $p_{i_2}$ and will grant $p_{i_1}$ more bonus points. 
    Their bonus points are both normalized by $\textrm{cosm}^{-1}(\textrm{btm},\textrm{top})$, which is commonly regarded as the largest skill gap. 
    In some situations, however, skill gap to the top player may even exceed $\textrm{cosm}^{-1}(\textrm{btm},\textrm{top})$, such as $\overline{\textrm{cosm}}^{-1}(i_3,\textrm{top})$. 
    As a result, $p_{i_3}$ is thought to have a lower skill level and will receive fewer bonus points. 
    }
    \label{fig:sim2top}
\end{figure}

We now define $\textrm{Sim}_\textrm{top}(i)$ as follows, which calculates the skill level similarity between $\bm{x}_i$ and $\bm{x}_\textrm{top}$: 
\begin{equation}
\label{eqn:sim2top}
    \textrm{Sim}_\textrm{top}(i) = \dfrac{\overline{\textrm{cosm}}(i,\textrm{top})}{\textrm{cosm}(\textrm{btm},\textrm{top})}\, .
\end{equation}
Equation \eqref{eqn:sim2top} can be interpreted from a skill gap perspective. 
Let $\textrm{Sim}_\textrm{top}(i)=\left( {\overline{\textrm{cosm}}^{-1}(i,\textrm{top})}/{\textrm{cosm}^{-1}(\textrm{btm},\textrm{top})} \right)^{-1}$. 
Function $\overline{\textrm{cosm}}^{-1}(i,\textrm{top})$ represents the skill gap to the top player for $p_i$. 
\autoref{fig:sim2top} is a detailed explanation of this. 

Finally, we will grant the active players (including the top player and the bottom player) bonus points, scaled by $\textrm{Sim}_\textrm{top}(i)$: 
\begin{equation*}
    \hat{r_i} = r_i + k \times \textrm{Sim}_\textrm{top}(i),
 \end{equation*}
where $k$ is the same K-factor previously assigned for Elo. 
The larger the $\textrm{Sim}_\textrm{top}(i)$ is, the smaller the skill gap to the top player for $p_i$, and the more bonus points $p_i$ will get. 
To avoid score inflation, one could re-center the scores of all players.


\section{Experiments}

\subsection{Experimental Setting}

The parameters for our experiments are listed as follows: 
The K-factor is $k = 50$. 
For the random walk procedure of GElo, $16$ random walk sequences will be generated starting from each node in the graph, and the length of each sequence is $100$. 
For the embedding procedure of GElo, the context size that forms the neighborhood is set as $u=5$, and the dimension of vector embeddings is $300$. 

Experiments are conducted on a 3.2GHz eight-core AMD Zen~2 machine with 32GB memory. 
The major part of GElo is written in Julia, a high-performance scientific programming language. 
An open-source framework Gensim~\cite{rehurek2011gensim} is used for vector space modeling, with parameters $epoches=5$ and $batch\_words=10000$. 

Public datasets from online video games and sports leagues are adopted in our experiments, including both solo and team competition scenes. 
Following are introductions to the experiment datasets:
\begin{itemize}

    \item SC2\footnote{\url{https://github.com/alimbekovKZ/starcraft2-matches-history-predict}.}: This dataset is from Starcraft~II, a real-time strategy game. 
    It includes professional player matches from 2016 to 2017. 
    The match form is 1v1 competition. 
   
    \item ATP\footnote{\url{https://github.com/JeffSackmann/tennis_atp}.}: This dataset includes solo tennis matches organized by the Association of Tennis Professionals from 2003 to 2016. 
    The match form is 1v1 competition. 

    \item CS\_1\&CS\_2\footnote{\url{https://www.kaggle.com/mateusdmachado/csgo-professional-matches}.}: 
    These two datasets collect professional match histories from Counter-Strike: Global Offensive, a first-person shooter game. 
    CS\_1 includes matches from 2015 to 2016.
    CS\_2 includes matches from 2019 to 2020. 
    CS\_2 has a bigger and steadier player base than CS\_1. 
    The match form is 5v5 team competition. 

    \item LOL~\cite{gong2020optmatch}: This dataset is from League of Legends, a multiplayer online battle arena game. 
    It includes professional championship matches. 
    The match form is 5v5 team competition. 
    
\end{itemize}

\begin{table*}[!ht]

    \centering
    \caption{Prediction Error Rate}
    \label{tab:summary}

    \begin{tabular}{cccccccccccccccc} 
\toprule
\multirow{2}{*}{} & \multirow{2}{*}{} & \multicolumn{2}{c}{SC2} & \multicolumn{1}{l}{} & \multicolumn{2}{c}{ATP} & \multicolumn{1}{l}{} & \multicolumn{2}{c}{CS\_1} & \multicolumn{1}{l}{} & \multicolumn{2}{c}{CS\_2} & \multicolumn{1}{l}{} & \multicolumn{2}{c}{LOL} \\ 
\cmidrule(l){3-4}\cmidrule(l){6-7}\cmidrule(l){9-10}\cmidrule(l){12-13}\cmidrule(l){15-16}
 &  & Elo & GElo &  & Elo & GElo &  & Elo & GElo &  & Elo & GElo &  & Elo & GElo \\ 
\midrule
Sub-dataset: &  &  &  &  &  &  &  &  &  &  &  &  &  &  &  \\
1 &  & 0.2790 & \textbf{0.2694} &  & 0.3422 & \textbf{0.3388} &  & 0.3503 & \textbf{0.3455} &  & 0.3826 & \textbf{0.3716} &  & 0.3797 & \textbf{0.3796} \\
2 &  & 0.2707 & \textbf{0.2642} &  & 0.3357 & \textbf{0.3326} &  & \textbf{0.3439} & 0.3460 &  & 0.3673 & \textbf{0.3533} &  & \textbf{0.3849} & 0.3884 \\
3 &  & 0.2594 & \textbf{0.2446} &  & 0.3326 & \textbf{0.3310} &  & 0.3887 & \textbf{0.3661} &  & 0.3847 & \textbf{0.3473} &  & 0.3900 & \textbf{0.3862} \\
4 &  & 0.2651 & \textbf{0.2549} &  & 0.3333 & \textbf{0.3253} &  & 0.3918 & \textbf{0.3776} &  & 0.4284 & \textbf{0.4009} &  & \textbf{0.3878} & 0.3895 \\
5 &  & 0.2607 & \textbf{0.2527} &  & 0.3297 & \textbf{0.3249} &  & 0.3811 & \textbf{0.3558} &  & 0.4020 & \textbf{0.3753} &  & 0.3897 & \textbf{0.3896} \\
6 &  & 0.2485 & \textbf{0.2400} &  & 0.3307 & \textbf{0.3216} &  & 0.3607 & \textbf{0.3473} &  & 0.3818 & \textbf{0.3668} &  & \textbf{0.3771} & 0.3782 \\
7 &  & 0.2561 & \textbf{0.2517} &  & 0.3207 & \textbf{0.3148} &  & 0.3628 & \textbf{0.3445} &  & 0.3847 & \textbf{0.3779} &  & \textbf{0.3533} & 0.3573 \\
8 &  & 0.2651 & \textbf{0.2534} &  & 0.3310 & \textbf{0.3193} &  & \textbf{0.3725} & 0.3760 &  & 0.4107 & \textbf{0.4090} &  & 0.3530 & \textbf{0.3491} \\
9 &  & 0.2662 & \textbf{0.2597} &  & 0.3368 & \textbf{0.3326} &  & 0.3958 & \textbf{0.3782} &  & 0.4069 & \textbf{0.3915} &  & \textbf{0.3858} & 0.3873 \\
10 &  & 0.2630 & \textbf{0.2536} &  & 0.3261 & \textbf{0.3216} &  & 0.4061 & \textbf{0.3899} &  & \textbf{0.3803} & 0.3808 &  & \textbf{0.3930} & 0.3977 \\ 
\midrule
Summary: &  &  &  &  &  &  &  &  &  &  &  &  &  &  &  \\
Avg. &  & 0.2634 & \textbf{0.2544} &  & 0.3319 & \textbf{0.3263} &  & 0.3754 & \textbf{0.3627} &  & 0.3929 & \textbf{0.3774} &  & \textbf{0.3794} & 0.3803 \\
Lower CI &  & 0.2575 & \textbf{0.2483} &  & 0.3277 & \textbf{0.3210} &  & 0.3606 & \textbf{0.3506} &  & 0.3798 & \textbf{0.3636} &  & \textbf{0.3690} & 0.3693 \\
Upper CI &  & 0.2692 & \textbf{0.2606} &  & 0.3361 & \textbf{0.3315} &  & 0.3902 & \textbf{0.3748} &  & 0.4061 & \textbf{0.3913} &  & \textbf{0.3899} & 0.3913 \\
$t$-statistic &  & \multicolumn{2}{c}{9.6767} &  & \multicolumn{2}{c}{5.7846} &  & \multicolumn{2}{c}{4.0762} &  & \multicolumn{2}{c}{4.0829} &  & \multicolumn{2}{c}{-0.9032} \\
$P$-value &  & \multicolumn{2}{c}{$<$0.001***} &  & \multicolumn{2}{c}{$<$0.001***} &  & \multicolumn{2}{c}{0.0028**} &  & \multicolumn{2}{c}{0.0027**} &  & \multicolumn{2}{c}{0.3900} \\ 
\bottomrule
\multicolumn{16}{l}{
Significance level for $t$-test is $0.05$; ***, ** refer to significance levels of $P<0.001$, $P<0.01$. 
} \\
\multicolumn{16}{l}{
Each sub-dataset result is the average of five simulations. 
}
\end{tabular}

\end{table*}

\subsection{Prediction Error Rate}

Prediction error rate is the key metric for the skill evaluation performance of a rating system. 
It is calculated as the fraction of incorrectly predicted matches. 
A match that the higher-rated player loses is counted as an incorrectly predicted match. 
Analogous metrics are used in other studies~\cite{herbrich2006trueskill, ebtekar2021elo}. 

We generate several sub-datasets from each dataset to fully test GElo. 
The datasets are sorted by date in ascending order. 
The time unit is one year for ATP and one month for the rest. 
Each dataset is split into $13$ time units of data, marked as $t_1$, $t_2$, $t_3$, $\dots$, $t_{13}$ in order. 
Then, we make these $13$ time units of data into $10$ sub-datasets, including match records in $[t_1, t_2, t_3, t_4]$, $[t_2, t_3, t_4, t_5]$, $[t_3, t_4, t_5, t_6]$, $\dots$, $[t_{10}, t_{11}, t_{12}, t_{13}]$. 
For a sub-dataset with match records in $[t_1, t_2, t_3, t_4]$, we use both Elo and GElo to calculate their respective rating scores of players from the match records in $[t_1, t_2]$, and examine the prediction error rate in $[t_3, t_4]$ using Elo. 
For new players added in $[t_3, t_4]$, we set their default rating scores as the average final score of all players from $[t_1, t_2]$. 
Similarly for other sub-datasets. 

We present detailed experimental results in \autoref{tab:summary}. 
The upper part of the table includes the prediction error rates of every sub-dataset. 
We repeat simulations on each sub-dataset five times and keep the average results as the final results. 
The lower part of \autoref{tab:summary} reports summary metrics, including the average prediction error rate, the 95\% confidence intervals, and two-tailed paired $t$-test analyses that measure the difference between the corresponding $10$ sub-datasets. 
Bold numbers indicate better results.

The results show that GElo outperforms Elo in four of five datasets. 
GElo gets a consistently lower prediction error rate than Elo in all SC2 and ATP sub-datasets. 
The difference in prediction error rate is significant at the $0.001$ level. 
GElo also performs better in most CS\_1 and CS\_2 sub-datasets.  
The difference in prediction error rate is significant at the $0.01$ level.
GElo only gets a worse result in LOL dataset, but the difference is not significant. 

In comparison to Elo, GElo gets significantly lower prediction error rates in SC2, ATP, CS\_1, and CS\_2 datasets. 
It obtains excellent results for 1v1 competitions and good results for team vs. team competitions. 
Subsequent experiments are all conducted using sub-datasets generated in this subsection. 

\begin{figure}[!t]
    \centering
    \includegraphics[width=7.1cm]{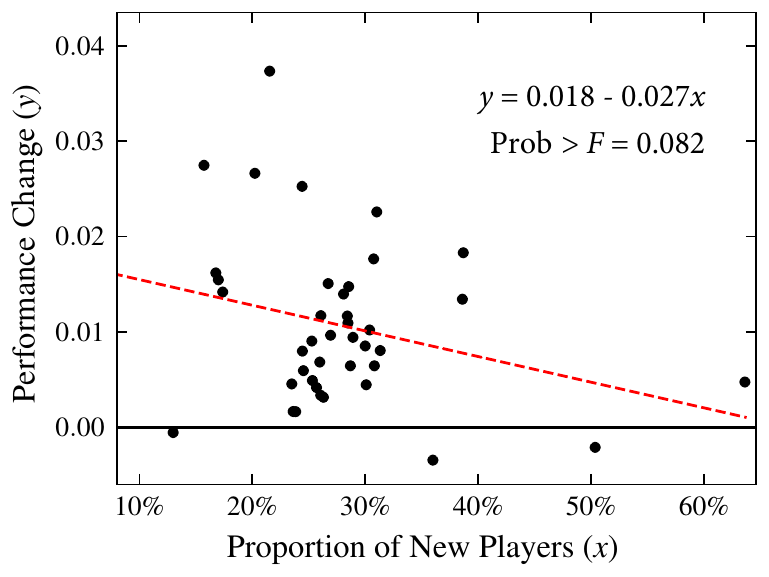}
    \caption{
    Relation between performance changes and proportion of new players. 
    Black dots represent the results from the sub-datasets of SC2, ATP, CS\_1, and CS\_2 in \autoref{tab:summary}. 
    "Performance Changes ($y$)" is calculated as the prediction error rate of Elo minus the one of GElo, and the larger the better. 
    The dashed red line is a linear regression line of $y$ on $x$. 
    }
    \label{fig:lm}
\end{figure}

\subsection{Performance Change with Increasing New Players}

This subsection examines how the prediction performance changes when there are increasing new players. 
For a sub-dataset with match records in $[t_1, t_2, t_3, t_4]$, players from $[t_3, t_4]$ without match records in $[t_1, t_2]$ are regarded as new players. 
Competitive games, especially the popular online video games nowadays, gain new players round the clock. 
For this reason, the performance change of a rating system when facing increasing new players is also an important metric. 

\autoref{fig:lm} visualizes the relationship between ``Performance Change ($y$)'' and ``Proportion of New Players ($x$)''. 
Black dots represent the results from the $40$ sub-datasets of SC2, ATP, CS\_1, and CS\_2 in \autoref{tab:summary}. 
$y$ is calculated as the prediction error rate of Elo minus the one of GElo, the larger the better, and $x$ is calculated as the newly-added players from $[t_3, t_4]$ divided by the total player amounts in $[t_1, t_2, t_3, t_4]$. 
The dashed red line is a linear regression model. 
The coefficient of $x$ on $y$ is $-0.027$ with $P\textrm{-value} = 0.082$. 

As the proportion of new players increases, the performance change of GElo against Elo reduces. 
It shows a slightly worse prediction performance for increasing new players under the condition that the proportion of new players is lower than $50\%$, but the change is not significant.

\subsection{Impact on Player Rank}

\begin{table}[!t]
    \centering
    \caption{Rank Variation (RV) for a Player}
    \label{tab:rv}
    
    \begin{tabular}{crrrrrr} 
\toprule
 & \multicolumn{3}{c}{(a) Avg. RV Per Match} & \multicolumn{3}{c}{(b) Avg. RV Per Sub-dataset} \\ 
\cmidrule(r){2-4}\cmidrule(r){5-7}
Dataset & \multicolumn{1}{c}{Elo} & \multicolumn{1}{c}{GElo} & \multicolumn{1}{c}{Change} & \multicolumn{1}{c}{Elo} & \multicolumn{1}{c}{GElo} & \multicolumn{1}{c}{Change} \\ 
\midrule
SC2 & 51.17 & 33.30 & -34.92\% & 326.38 & 259.39 & -20.53\% \\
ATP & 40.86 & 29.01 & -29.00\% & 198.40 & 160.19 & -19.26\% \\
CS\_1 & 92.24 & 78.84 & -14.52\% & 253.72 & 230.92 & -8.99\% \\
CS\_2 & 125.85 & 94.70 & -24.75\% & 307.83 & 248.85 & -19.16\% \\
\bottomrule
\end{tabular}

\end{table}

In this subsection, we will observe the rank variation (RV) for a player. 
Suppose a player initially has the 15th highest rating score of all players, and after a certain period, this player has the 10th (or 20th) highest rating score. 
Then, this player is considered to have a rank variation of $5$.

\autoref{tab:rv} compares the average rank variation for a player between Elo and GElo in SC2, ATP. CS\_1, and CS\_2 datasets. 
Test (a) is "Avg. RV Per Match", the average RV for a player after the player finishes a match. 
Test (b) is "Avg. RV Per Sub-dataset". 
For a sub-dataset with match records in $[t_1, t_2, t_3, t_4]$, test (b) compares the average rank difference for a player between $t_2$ and $t_4$. 

The results show that GElo has a smaller rank variation than Elo. 
Compared to vanilla Elo, the rank variation of GElo is, on average, 25\% lower in test~(a) and 17\% lower in test~(b). 
Not many players want their ranks to vary frequently. 
GElo establishes a steadier rank system compared to vanilla Elo.

\subsection{Running Time Analysis}

\begin{table}
    \centering
    \caption{Running Time Analysis}
    \label{tab:running_time}
    
        \begin{tabular}{crrrrr} 
    \toprule
     & \multicolumn{3}{c}{Dataset Information} & \multicolumn{2}{c}{Time Cost (seconds)} \\ 
    \cmidrule(r){2-4}\cmidrule(r){5-6}
    \multicolumn{1}{r}{Dataset} & \textbar{}Nodes\textbar{} & \textbar{}Edges\textbar{} & \textbar{}Matches\textbar{} & Walking & Embedding \\ 
    \midrule
    SC2 & 1008~ & 4227 & 8393~~ & 1.52~~ & 14.42~~~ \\
    ATP & 640~ & 4887 & 6273~~ & 1.56~~ & 7.81~~~ \\
    CS\_1 & 721~ & 10592 & 606~~ & 1.87~~ & 10.47~~~ \\
    CS\_2 & 1158~ & 21148 & 1049~~ & 3.83~~ & 17.09~~~ \\
    LOL & 531~ & 12238 & 1713~~ & 1.20~~ & 9.01~~~ \\
    \bottomrule
    \multicolumn{6}{l}{Each ``Time Cost'' result is the average of five simulations.} \\
    \end{tabular}

\end{table}

\begin{figure}[!t]
    \centering
    \subfloat[SC2]{\includegraphics[width=4.25cm, frame]{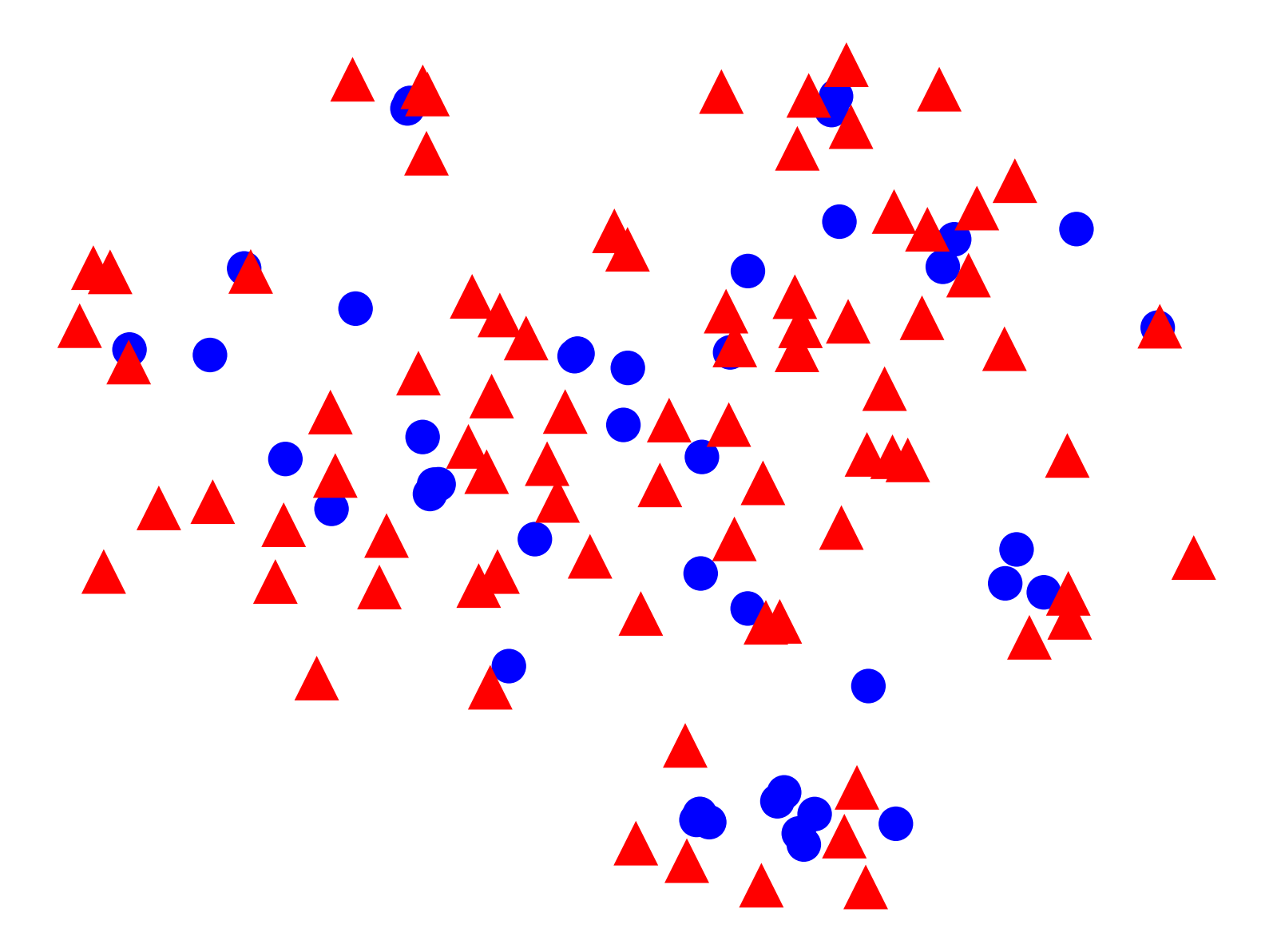}} \hfil
    \subfloat[ATP]{\includegraphics[width=4.25cm, frame]{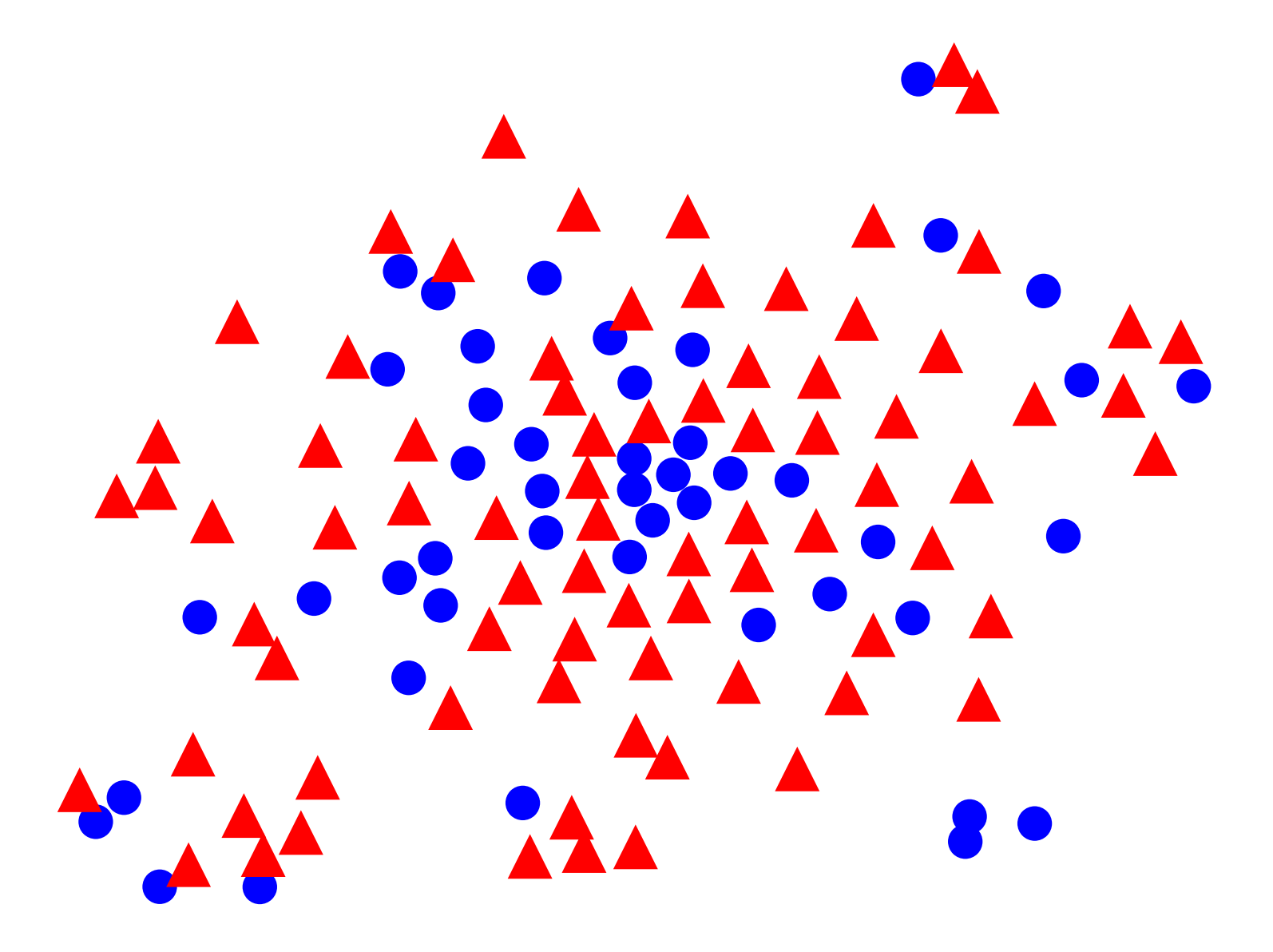}} \\
    \subfloat[CS\_1]{\includegraphics[width=4.25cm, frame]{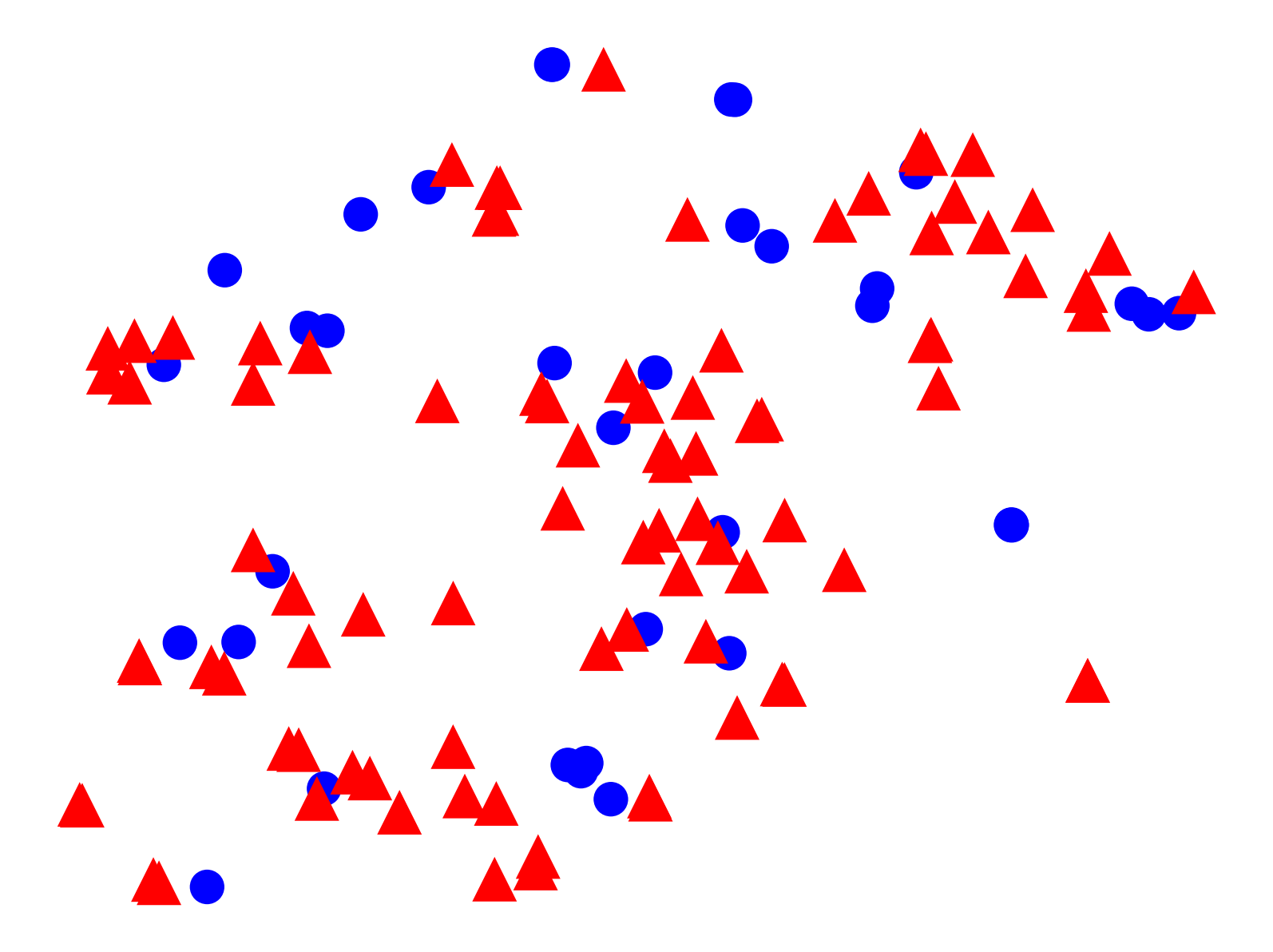}} \hfil
    \subfloat[CS\_2]{\includegraphics[width=4.25cm, frame]{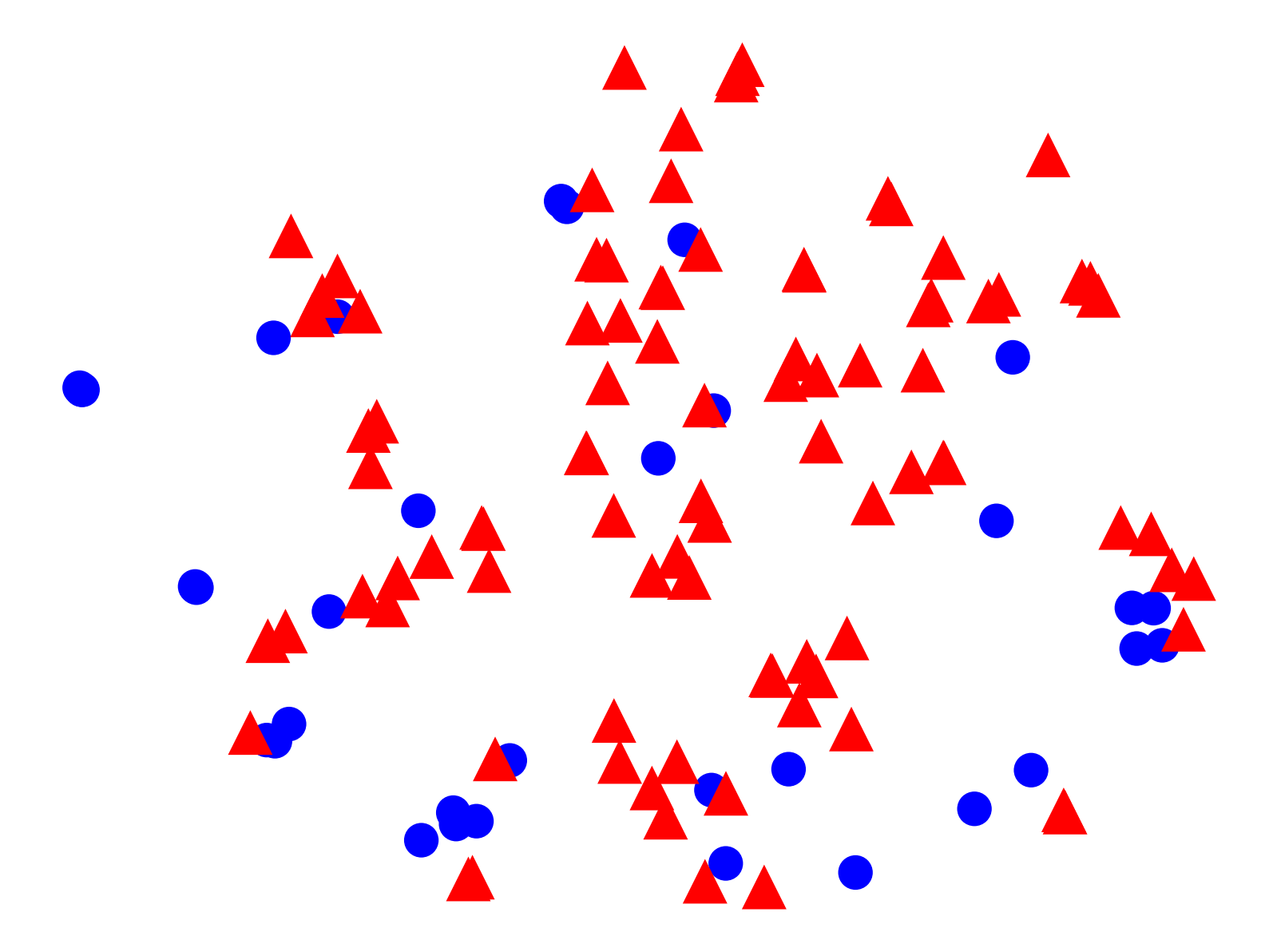}} \\
    \caption{
    Visualizations of player embeddings. 
    Red triangles represent active players, and blue dots represent inactive players. 
    It can be observed that red triangles spread across the whole vector space. 
    This means the active players cover most types of players. 
    }
    \label{fig:embvis}
\end{figure}

GElo generates random walk sequences and trains them using Skip-gram model to generate node embeddings. 
An empirical running time analysis is presented in~\autoref{tab:running_time}. 
The program is executed using $16$ simultaneous threads. 
The results show a good efficiency of GElo.

\subsection{Visualization of Player Embeddings}

\autoref{fig:embvis} visualizes the player embeddings we learned from SC2, ATP, CS\_1, and CS\_2 through t-SNE method. 
Red triangles represent active players, while blue dots represent inactive players. 
From the spread pattern of the visualized vectors, we can observe that the embeddings of the active players are broadly distributed across the whole player group, meaning that those active players cover a diverse range of player styles and characteristics.


\section{Conclusion}

This paper sets out to present a framework that learns player embeddings by constructing a skill gap graph based on player match histories. 
These vectorized graph embeddings preserve the win-loss relationships among players and can be conveniently applied to subsequent tasks. 
Then, we introduce Graphical Elo (GElo), which incorporates player embeddings to improve the performance of evaluation player skills. 
The experiments show that GElo outperforms vanilla Elo in four of five real-world datasets. 
Our work suggests potential applications of player embeddings. 

It is unfortunate that the work does not include a rigorous theoretical framework for the skill gap graph and its interaction patterns. 
Another limitation lies in the fact that the current skill gap graph cannot well reflect the skill gap between a pair of players who only meet once. 
In spite of the existing limitations, the study certainly adds to our understanding of the skill rating process from a graphical perspective. 

Our application of the player embeddings, which are primarily used to scale the bonus points granted to the active players, is merely an example and far from the full potential. 
Further research can design new types of player graphs, employ different graph embedding methods, or apply player embeddings to various tasks in competitive games.


\section*{Acknowledgement}

The author would like to sincerely thank the reviewers and editors who kindly reviewed the earlier versions of the manuscript. 
Their valuable comments and suggestions have greatly improved the quality of this paper. 

\ifCLASSOPTIONcaptionsoff
  \newpage
\fi

\balance

\bibliography{refs}


\end{document}